\documentclass{article}

\usepackage{PRIMEarxiv}

\usepackage[utf8]{inputenc} 
\usepackage[T1]{fontenc}    
\usepackage{hyperref}       
\usepackage{url}            
\usepackage{booktabs}       
\usepackage{amsfonts}       
\usepackage{nicefrac}       
\usepackage{microtype}      
\usepackage{lipsum}
\usepackage{fancyhdr}       
\usepackage{graphicx}       
\graphicspath{{media/}}     

\usepackage{cite}
\usepackage{amsmath,amssymb,amsfonts}
\usepackage{algorithmic}
\usepackage{graphicx}
\usepackage{textcomp}
\usepackage{booktabs} 
\usepackage{tabularx} 
\usepackage{xcolor}
\usepackage{ulem}
\usepackage{url}
\usepackage{hyperref}

\usepackage{algorithm}
\usepackage{soul}
\usepackage{amsmath}
\usepackage{subcaption} 
 
\usepackage{amssymb}
\usepackage{pifont}
\newcommand{\cmark}{\ding{51}}%
\newcommand{\xmark}{\ding{55}}%

\title{VisionGPT: LLM-Assisted Real-Time Anomaly Detection for Safe Visual Navigation
}

\author{
  Hao Wang \\
  School of Computing \\
  Clemson University \\
  Clemson, SC, USA\\
  \texttt{hao9@g.clemson.edu} \\
   \And
  Jiayou Qin \\
  Department of Electrical and Computer Engineering \\
  Stevens Institute of Technology \\
  Hoboken, NJ, USA\\
  \texttt{jqin6@stevens.edu} \\
  \AND
  Ashish Bastola \\
  School of Computing \\
  Clemson University\\
  Clemson, SC, USA\\
  \texttt{abastol@g.clemson.edu } \\
  \And
  Xiwen Chen \\
  School of Computing \\
  Clemson University\\
  Clemson, SC, USA\\
  \texttt{xiwenc@g.clemson.edu}\\
  \And
  Zihao Gong \\
  School of Cultural and Social Studies \\
  Tokai University\\
  Tokyo, Japan \\
  \texttt{0CPD1206@mail.u-tokai.ac.jp} \\
\And
  John Suchanek \\
  School of Computing \\
  Clemson University\\
  Clemson, SC, USA\\
  \texttt{jsuchan@g.clemson.edu } \\
\And
  Abolfazl Razi \\
  School of Computing \\
  Clemson University\\
  Clemson, SC, USA\\
  \texttt{arazi@clemson.edu } \\
}

\begin{document}
\maketitle

\begin{abstract}
This paper explores the potential of Large Language Models(LLMs) in zero-shot anomaly detection for safe visual navigation. With the assistance of the state-of-the-art real-time open-world object detection model Yolo-World and specialized prompts, the proposed framework can identify anomalies within camera-captured frames that include any possible obstacles, then generate concise, audio-delivered descriptions emphasizing abnormalities, assist in safe visual navigation in complex circumstances. 
Moreover, our proposed framework leverages the advantages of LLMs and the open-vocabulary object detection model to achieve the dynamic scenario switch, which allows users to transition smoothly from scene to scene, which addresses the limitation of traditional visual navigation.
Furthermore, this paper explored the performance contribution of different prompt components, provided the vision for future improvement in visual accessibility, and paved the way for LLMs in video anomaly detection and vision-language understanding. 

\end{abstract}


\section{Introduction}
Accessible technologies have seen remarkable development in recent years due to the rise of machine learning and mobile computing \cite{donner2015after, al2020smart, khan2020ai, li2018vision, bastola2023multi}. Deep learning has significantly enhanced the accuracy and speed of object detection and segmentation models \cite{afif2020evaluation, kacorri2017people, bhandari2021object, ashiq2022cnn, bastola2023multi}, which catalyzed a surge of real-world applications, impacting numerous aspects of daily life, industry, and transportation. Visual navigation has benefited significantly from the evolution of such computer vision techniques \cite{kuzdeuov2023chatgpt, bastola2023multi, bastola2024driving}. 

Consequently, innovations such as Augmented Reality (AR) have been instrumental in enhancing the safety and mobility of individuals across various scenarios, including driving and walking. Many of these technologies aim to bridge the gap between the physical world and digital assistance, highlighting the critical need for adaptive solutions to navigate the complexities of real-world environments.


However, visual navigation presents significant challenges in dynamic urban environments \cite{zhao2024vialm, bastola2024driving, bastola2023feedback}. 
Although the newborn zero-shot object detection \cite{bansal2018zero} addresses the significant limitations of classical object detection models such as YOLOv8 \cite{redmon2016you, Jocher_Ultralytics_YOLO_2023} in complex scenarios, it encounters difficulties in developing custom class labels for dynamic environments due to the long-tail response. 
Furthermore, real-time vision-language understanding can be critical for complex scenarios for safety concerns, especially for visually impaired individuals who must traverse streets, sidewalks, and other public spaces. 

Vision-language understanding has recently become a hotspot due to the emergence of Multimodal Large Language Models (LLMs) \cite{wu2024vision}. Multimodal LLMs represent an evolutionary leap in the field of artificial intelligence as they integrate the processing of text, images, and even audio and video \cite{achiam2023gpt} to create a comprehensive understanding of the world that mirrors human cognitive abilities more closely than ever before, making it possible to handle more advances tasks for robotics \cite{zhang2023interactive, li2023improving}. 
Specifically, GPT-4V is now being heavily used in image tasks such as data evaluation \cite{tian2024supervised, cao2023towards}, medical image diagnosis \cite{nori2023capabilities, yang2023performance, waisberg2023gpt}, and content creation \cite{sridhar2023harnessing, sun2023evaluating, gimpel2023unlocking}.

\begin{figure*}[ht]
    \centering
    \centerline{\includegraphics[width=1\textwidth]{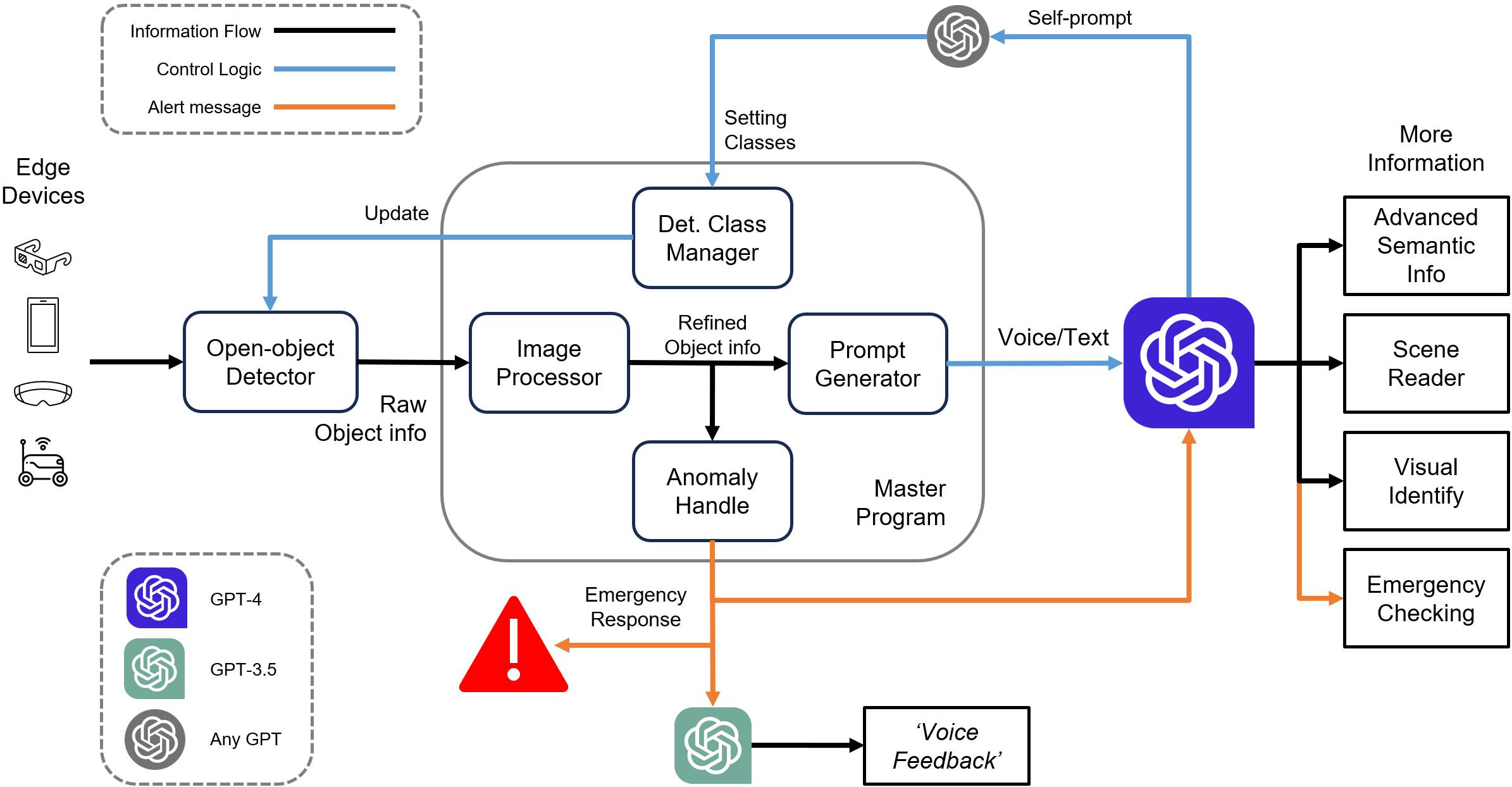}}
    \caption{Framework for vision-language processing and prompting.}
    \label{fig:frame}
\end{figure*}

Multimodal LLMs possess substantial improvement in interpreting, analyzing, and generating content across different modalities \cite{radford2021learning}, bringing the possibility to interdisciplinary applications \cite{luddecke2022image, liu2024visual}. 
Interestingly, LLMs also exhibit impressive zero-shot and few-shot learning abilities, potentially enabling them to capture visual concepts with minimal training data. This opens a way to address object detection challenges, particularly in data with limited annotation \cite{liao2024gpt}.
Recent research attempts to bring LLMs to the accessibility field, yet most work only focuses on basic natural language processing such as text reading, image recognition, and voice assistance \cite{yan2023gpt}. 


Therefore, a critical gap exists when using the vision-language understanding of LLMs for safety and accessible applications. Despite past works that investigated the use of LLMs in visual assistance \cite{chen2024mapgpt} and visual navigation \cite{pan2023langnav, shah2023lm, zhang2024navhint}, only a few focused on the safety aspects \cite{liao2024gpt, hwang2024safe} but barely considered the induced latency during the inference.

Our research introduces a framework that combines the speed of locally executed open-world object detection with the intelligence of LLMs to create a universal anomaly detection system. 
The primary goal of this system is to deliver real-time, personalized scene descriptions and safety notifications, ensuring the safety and ease of navigation for visually impaired users by identifying and alerting them to potential obstacles and hazards in their path, where these obstacles and hazards can be considered "anomalies" in the context of a safe and clear path for navigation. The proposed framework can also be applied to robotic systems, augmented reality platforms, and all other mobile computing edge units.


Particularly, the major contributions of this paper are summarized as follows:
\begin{itemize}
\item Zero-shot anomaly detection: The proposed integration is train-free and ready for video anomaly detection and annotation with different response preferences.
\item Real-time feedback: Our framework is optimized for real-time response in complex scenarios with very low latency.
\item Dynamic scene transition and interest setting: This framework can dynamically switch the object detection classes based on the user's needs. Furthermore, users can interact with the LLM module and setting a prior task (e.g., find the nearest bench). 
\end{itemize}


\section{Related Work}

\subsection{Open-vocabulary object detection}
Open-vocabulary object detection (OVD)\cite{zareian2021open} represents a significant shift in object detection focusing on identifying items outside predefined categories. Initial efforts\cite{gu2021open} trained on known classes for evaluating the detection of novel objects facing generalization and adaptability issues due to limited datasets and vocabularies. Recent approaches, however, \cite{zhong2022regionclip, zhou2022detecting, wu2023aligning} employ image-text matching with extensive data to expand the training vocabularies inspired by vision language pre-training\cite{ radford2021learning, jia2021scaling}. OWL-ViTs\cite{heigold2023video} and GLIP\cite{li2022grounded} utilize vision transformers and phrase grounding for effective OVD while Grounding DINO\cite{liu2023grounding} combines these with detection transformers for cross-modality fusion.  Despite the promise, existing methods often rely on complex detectors increasing computational demands significantly\cite{zhang2022dino, zhang2020bridging}. ZSD-YOLO\cite{xie2021zsd} also explored an open-vocabulary detection with YOLO using language model alignment; however, YOLO-world\cite{cheng2024yolo} presents a more efficient and real-time OVD solution aiming to be much more efficient with real-time inference using an effective pre-training strategy while still being highly generalizable.

\subsection{Prompt Engineering}
Prompt engineering has emerged as a critical technique for unlocking the capabilities of large language models (LLMs) \cite{gu2023systematic, chang2023survey, wang2023prompt, bastola2023llm} to various applications without finetuning on large datasets. This involves carefully crafting text prompts, instructions, or examples to guide LLM behavior and elicit desired responses. Researchers are actively exploring prompt engineering using various prompting techniques such as zero-shot prompting\cite{wan2023better}, Few-shot prompting\cite{brown2020language}, Chain-of-thought prompting\cite{wei2022chain}, self-ask prompting \cite{press2022measuring}, etc. to fine-tune LLMs for various tasks, demonstrating significant performance gains compared to traditional model training approaches. Studies have showcased how prompt engineering can adapt LLMs for diverse natural language tasks like question-answering\cite{zhuang2024toolqa}, smart-reply\cite{bastola2023llm}, summarization\cite{pilault2020extractive}, and text classification\cite{puri2019zero, clavie2023large}. Furthermore, researchers are increasingly developing frameworks to systematize prompt engineering efforts. Such frameworks aim to simplify the creation of effective prompts and facilitate the adaptation of LLMs to specific domains and applications and are highly customizable to user needs. While prompt engineering has seen significant improvements in natural language processing, its potential in computer vision on accessibility remains less explored. Our work builds upon the success of prompt engineering in NLP, exploring its application in the visual domain to enhance object detection and description.


\vspace{-0.05cm}

\subsection{Accessible Technology}
Computer vision-driven accessible technologies are mostly designed to empower individuals with visual impairments through enhanced scene understanding and hazard detection. A range of solutions exist, including mobile apps that provide object recognition and audio descriptions of surroundings\cite{joshi2020efficient, hakobyan2013mobile, matusiak2013object}, to wearable systems that offer real-time alerts about obstacles or potential dangers\cite{rahman2020improved, ramadhan2018wearable}. For example, technologies that detect approaching vehicles and crosswalk signals significantly improve the safety of visually impaired pedestrians in urban environments. Moreover, computer vision is integrated into assistive technologies for reading text aloud from documents and identifying objects in daily life, enabling greater independence\cite{bastola2023multi}. Research in this domain also focuses on indoor navigation, where object detection and spatial mapping can guide users within buildings and public spaces\cite{fallah2012user}.
The core emphasis of these computer vision-powered accessibility technologies aims to enhance safety. By providing real-time information on key elements within an individual's surroundings, the risk of accidents and injuries is significantly reduced. Identifying potential hazards, such as oncoming traffic, obstacles on sidewalks, or unattended objects, allows visually impaired individuals to navigate with greater confidence and autonomy.

\section{Methodology}
Our system offers real-time anomaly alerts by integrating object detection with large language model capabilities, featuring a multi-module architecture. 
The system operates continuously with the object detection module processing real-time camera frames.
Multi-frame object information is then included in specially engineered prompts and submitted to the LLM module. The system then processes the LLM's response, classifying potential anomalies. 
Finally, the LLM module conveys important alerts and essential scene descriptions to the user.

The proposed project is fully open-sourced and available at: \href{https://github.com/AIS-Clemson/VisionGPT}{https://github.com/AIS-Clemson/VisionGPT}

\subsection{Object Detection Module}
To ensure real-time performance on mobile devices, we employ lightweight yet powerful object detection models for real-time detection. 
Specifically, we applied the state-of-the-art YOLO-World model for open-vocabulary detection whose detection classes are customizable for a wide range of scenarios. As we focus on accessible visual navigation, we prompt the  proposed LLM module to personalize the detection classes relevant to safe navigation in daily use circumstances, including pedestrians, vehicles, bicycles, traffic signals, and any potential road hazards or obstacles. 
Therefore, our proposed multi-functional prompt manager allows users to switch detection classes dynamically. 


\subsection{Detection Class Manager}
This sub-module aims to create a detailed categorization for object detection algorithms, enabling them to identify and distinguish potential obstacles, hazards, and useful landmarks. 
This approach ensures the detection system is finely tuned to urban navigation's specific needs and challenges, enhancing the user's ability to move safely and independently through city streets. 
By focusing on road hazards and obstacles, the updated list aims to provide a more relevant and focused set of detection classes for the 'urban walking' context, optimizing the system's utility and effectiveness for the visually impaired user.

As shown in Figure \ref{fig:frame}, the user can interact with the LLM (advance) module. Based on this operation logic, the user can ask to change the object detection classes based on scenarios. 
For instance, if a user experiences a scene transition from sidewalk to park, the detection classes specialized for sidewalk objects (e.g., car, road cone, traffic signal, etc.) can be replaced by new object classes that are more relevant to the park scene to adapt to the situation.

Original prompt: \textit{"The user is switching the scene to {custom\_scene} please generate a new list that contains the top 100 related objects, including especially road hazards and possible obstacles"}




\subsection{Anomaly Handle Module}


 
The proposed anomaly detection system aims to enhance navigation safety and awareness in various environments, particularly for visually impaired individuals and others requiring navigation assistance (e.g., robotics systems). 
The system analyzes real-time imagery captured from a camera and splits the image into four distinct regions based on an 'H' pattern.

\begin{figure}[ht]
    \centering
    \centerline{\includegraphics[width=0.6\linewidth]{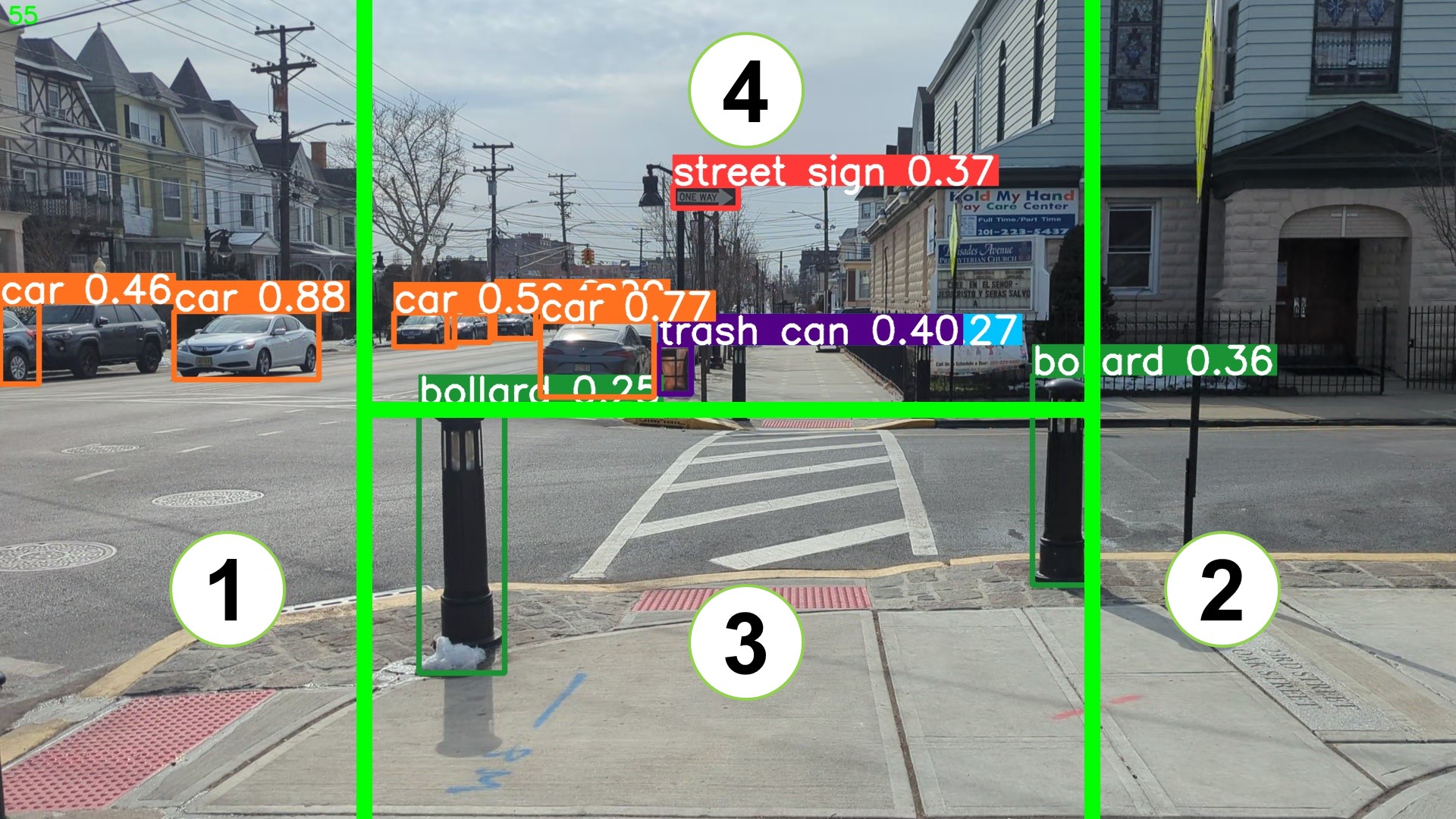}}
    \caption{Type H image splitter. (1) and (2) represent the left and right area, (3) represent the ground area, and (4) represent the front area.}
    \label{fig:H}
\end{figure}

Specifically, The system categorizes detected objects into four types based on their location within the image, each corresponding to a specific splitting part of the 'H' pattern segmentation: \textbf{Left}, \textbf{Right}, \textbf{Front}, and \textbf{Ground}. This categorization helps identify and respond to potential hazards more effectively. As shown in Figure \ref{fig:H}. 

Moreover, we find that:
\begin{itemize}
    \item \textbf{Left} and \textbf{Right}: These regions cover the left and right 25\% of the image, respectively. Objects detected here are typically in motion and may occupy much of the visual field. This area is crucial for identifying moving hazards such as vehicles or cyclists that may approach the user from the sides.
    \item \textbf{Front}: This region focuses on the center 50\% of the image's width and the upper half vertically. It captures objects still at a distance but directly ahead of the user. Identifying objects in this region is necessary for assessing the overall situation and planning movements, especially in detecting upcoming objects at high speed such as cars and cyclists.
    \item \textbf{Ground}: Occupying the center 50\% of the image's width and the lower half vertically, this area highlights objects nearby on the ground. Immediate attention to detections in this area is critical for avoiding hazards that require cautious navigation, such as cracks, puddles, or uneven surfaces.
\end{itemize}


The system then records detailed information for each object, including classification, size, and position. All size and position data have been converted into percentage expressions for a better interpretation by LLM.
Finally, by analyzing objects' locations and sizes, alerts for anomalies are generated for objects that appear on the 'ground' area or occupy significant space ($>10\%$ in this study) in the 'left' or 'right' regions.
The detection and movement information is then post-processed into a structured format, supporting LLM for better understanding. 

Original prompt: \textit{"The location information \(center_x, center_y, height, width\) of objects is the proportion to the image, the detected objects are categorized into 4 type based on the image region. Left and Right: objects located on left 25\% or right 25\% of the image, these objects are usually moving and has large proportion.Front: objects that may still far away, can be used to discriminate the current situation.Ground: objects that may nearby."}

\subsection{Data Collection}
Even though various datasets exist for static images\cite{tang2023dataset} and CCTV camera feeds\cite{li2012incremental, mehran2009abnormal, kim2022anomaly}, no extensive datasets are available for detecting large anomalies in visual navigation from the first person's perspective. 

Thus, we collected 50 video clips of point-of-view cruising in various scenarios.
These custom videos are filmed in public spaces with first-person view and continuous forward-moving. Table \ref{tab:data} shows the details of the collected data. 

\begin{table}[htbp]  
    \centering      
        \resizebox{0.9\linewidth}{!}{
    \label{tab:data}
    \begin{tabular}{cccccccc} 
        \toprule      
        Location&Scene  &Movement &Weather& Clips&Total length& Unique Classes& Total detected objects\\\midrule
 Urban  &Sidewalk&Scooter &Cloudy&   8&10 mins& 31& 16944\\
 Suburban& Bikeline& Scooter & Cloudy& 5& 6 mins& 26&8394\\ 
 Urban &Park&Scooter&Cloudy&   6&5 mins& 23&  15310\\
 City &Road& Biking&Sunny&   5&5 mins& 21&  5464\\
 City& Sidewalk& Biking& Sunny& 7& 6 mins& 27&9569\\
 City&Park& Biking&Cloudy&   5&5 mins& 19&  4781\\
 Town&Park& Walking&Cloudy&    6&4 mins& 18&  5156\\
 Town & Sidewalk& Walking & Sunny& 8& 7 mins& 14&8274\\
 City& Coast& Walking & Sunny& 2& 5 mins& 37&29280\\
        Suburban&Theme Park& Walking &Rain&   3&6 mins& 34&  24180\\ \bottomrule
    \end{tabular}}
    \caption{Collected data for video anomaly detection.}
\end{table}

We then conducted the experiments by combining the open-vocabulary object detection model with our novel image-splitting method to annotate the frames as anomalies. 

Specifically, a frame is labeled as an anomaly if it meets either of the following criteria:
\begin{enumerate}
    \item Objects are detected within the \textbf{Ground} area.
    \item Objects appear in either the \textbf{Left} or the \textbf{Right} areas of the image and occupy more than 10\% of the total image area.
\end{enumerate}

We set this rule-based method as the baseline of anomaly detection in this study, as our captured video clips are customized for this H-splitting principle. 


\subsection{LLM Module}
This module processes the detected object information and passing to the LLM. Specifically, we use both GPT-3.5 and GPT-4 to process the information. First, GPT-3.5 is mainly used for low-level information processes such as object detection data analysis, data format converting, and prompt reasoning. Therefore, GPT-4 is used for a high-level command instance understanding and a comprehensive vision-language understanding.

The system sensitivity settings indicate a focus on identifying and reporting hazards based on their potential impact on the user's safety and navigation. The system's goal to report objects based on their level of inconvenience or danger aligns with the anomaly detection objective of identifying and reacting to deviations that matter most in the given context. Note that the system sensitivity in the context is distinct from the model sensitivity as a statistical term.

These prompts sketch the conceptual framework and operational guidelines for a voice-assisted navigation system for visual accessibility. The system utilizes data from a phone camera, which is always facing forward, to detect objects and categorize their location within the field of view. Based on these analyses, the system provides auditory feedback to users, helping them navigate their environment safely and avoid potential hazards. Furthermore, the annotated data can be used for the training of other anomaly detection models.


The main LLM prompts consist of: 
\begin{itemize}
    \item Prompt instruction: \textit{"You are a voice assistant for a visually impaired user, \
the input is the actual data collected by a phone camera, and the phone is always facing front, \
please provide the key information for the blind user to help him navigate and avoid potential danger. \
Please note that the center\_x and center\_y represent the object location (proportional to the image), \
object height and width are also a proportion."}
    \item Prompt sensitivity: \textit{"System sensitivity: Incorporate the sensitivity setting in your response. \
For a low-sensitivity setting, identify and report only imminent and direct threats to safety. \
For medium sensitivity, include potential hazards that could pose a risk if not avoided. \
For high sensitivity, report all detected objects that could cause any inconvenience or danger.\
Current sensitivity: low."}
\end{itemize}

\section{Experiments}
We compare our proposed vision-LLM system with the rule-based anomaly detection (baseline) to show its performance and reliability.

\subsection{System Optimization}
While the proposed system is running, we input captured images into the object detection model every $\textbf{5}$ frames to boost the FPS (Frame-Per-Second), this can significantly improve the performance, especially for mobile devices that have limited computation resources. Then, we send the detected information to the anomaly handle module to label the frames as the baseline. With the frame compensation, the real-time detection performance is boosted from $\textbf{16}$ \textbf{FPS} to $\textbf{73}$ \textbf{FPS}, as shown in Table \ref{table:fps}. 

Meanwhile, we apply the LLM module to process the detected information every $\textbf{30}$ frames in parallel due to the latency of LLMs. To optimize the latency for better performance, the proposed system uses the GPT 3.5 Turbo model as the core of the LLM module. 

\begin{figure}[htbp]
    \centering
    \begin{minipage}[t]{0.5\textwidth}
        \includegraphics[width=\textwidth]{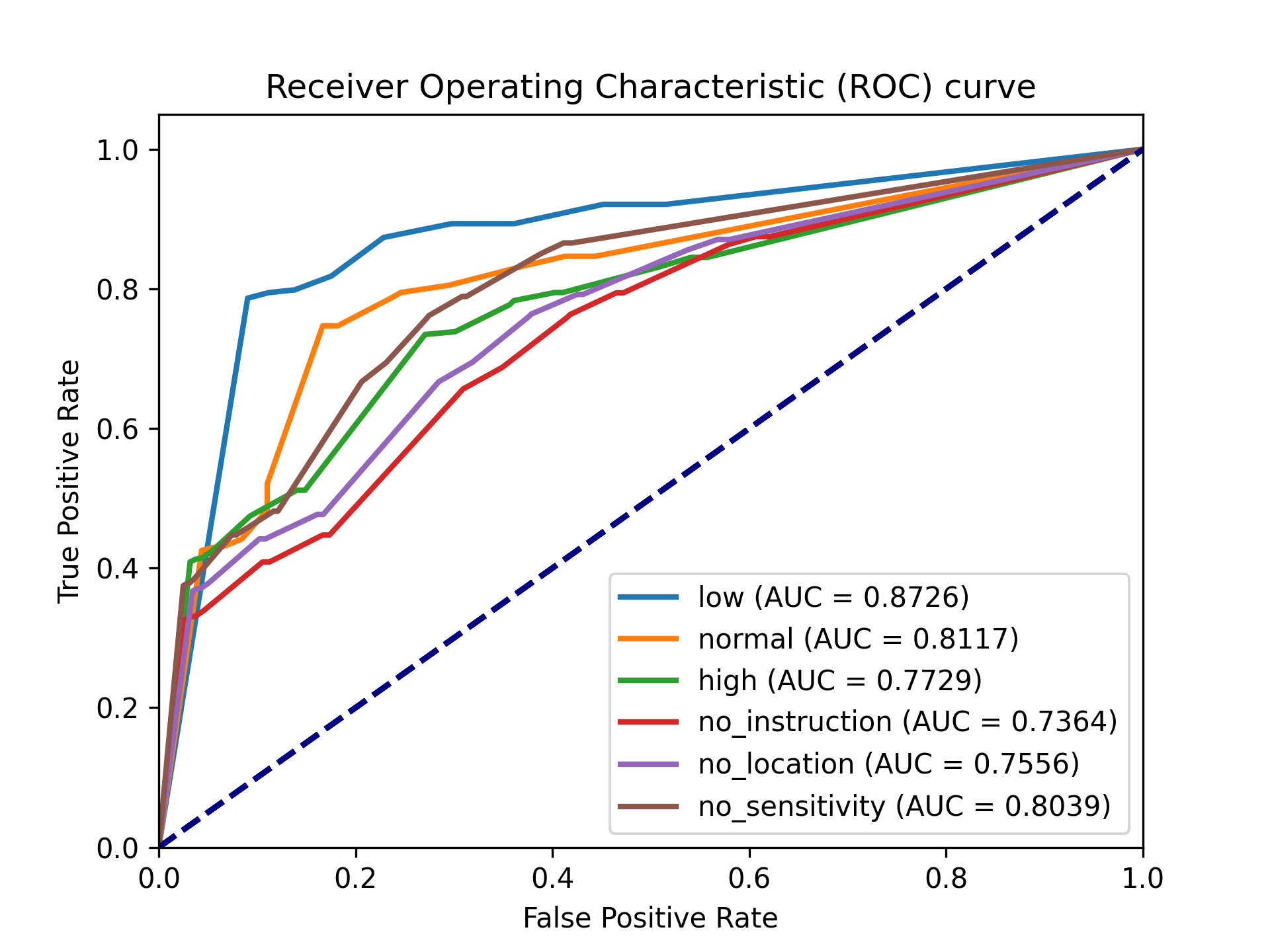}
        \caption{ROC curve.}
        \label{fig:roc}
    \end{minipage}
    \hfill
    \begin{minipage}[t]{0.4\textwidth}
        \includegraphics[width=\textwidth]{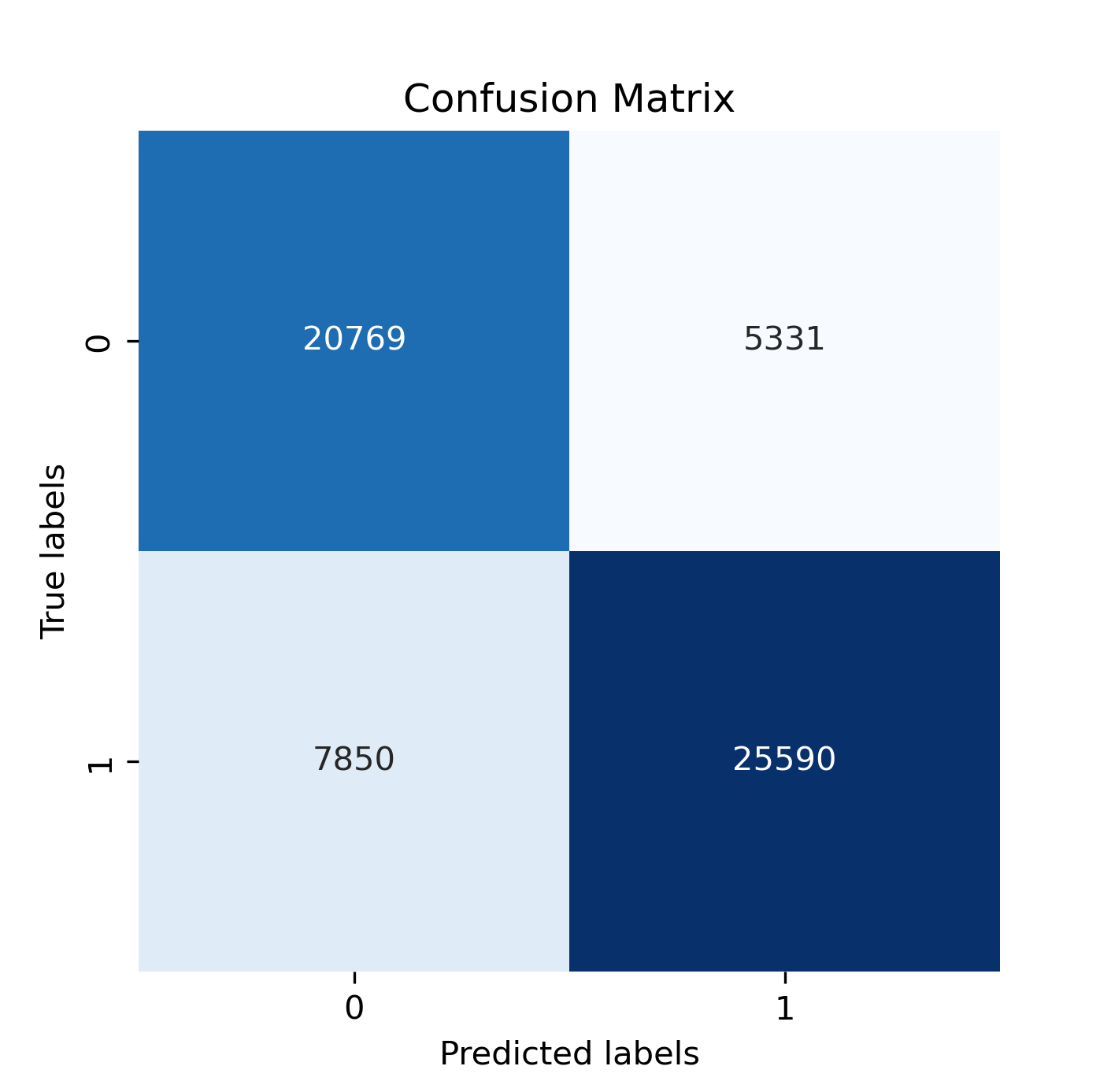}
        \caption{Confusion matrix of total frames. LLM setting is low-system sensitivity setting.}
        \label{fig:conf}
    \end{minipage}
\end{figure}



\subsection{Detection Accuracy}
By setting the rule-based detector as the baseline, this study aims to test the zero-shot learning capability of the LLM detector, and meanwhile, interpret which prompt may impact the performance significantly. 

After comparing the annotation results of prompt-based anomaly detection with the rule-based anomaly detection on our collected data, we find that prompt-based anomaly detection achieves high precision with all prompt modules working properly. 
Specifically, we compared the LLM anomaly detection with different sensitivity settings: low, normal, and high. As shown in Figure \ref{fig:roc}, the Receiver Operating Characteristic (ROC) curve indicates that a low system sensitivity leads to better performance, as it is less sensitive than the rule-based detector. 
For instance, objects detected by the rule-based detector with low confidence and classes of low risk will be filtered by LLM due to no emergency.
Conversely, the higher the system sensitivity, the worse the performance, as the system tends to categorize all possible anomalies as immediate emergencies.

As shown in Figure \ref{fig:conf}, the LLM anomaly detector with low-system sensitivity captures more True Positive and True Negative cases and tries to minimize the False Positive rate.





\subsection{Quality Evaluation}
\begin{figure}[ht]
    \centering
    \centerline{\includegraphics[width=0.8\linewidth]{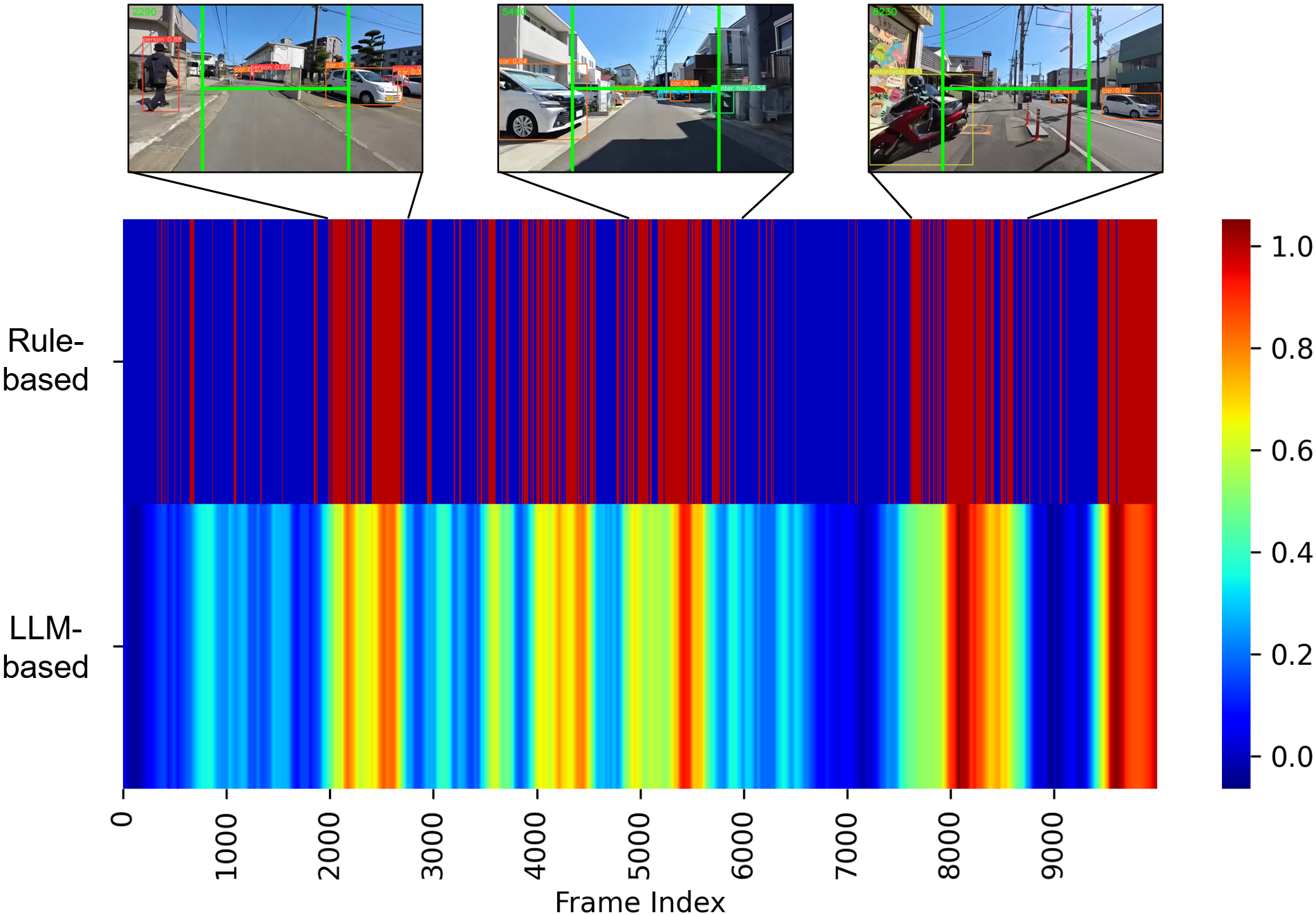}}
    \caption{Anomaly annotation. The first row represents the labeled anomalies by the rule-based detector (binary), and the second row represents the anomalies predicted by the proposed LLM detector (float). Color represents the probability of anomalies.}
    \label{fig:heatmap}
\end{figure}

We picked one of the video clips to analyze the detection difference between the rule-based detector and the LLM anomaly detector. Specifically, in Figure \ref{fig:heatmap}, the first row shows the anomalies labeled by the rule-based detector, while the second row indicates the anomalies predicted by the LLM-based detector (low sensitivity setting). As shown in Figure \ref{fig:heatmap}, the proposed LLM detector has less acuity with a low-sensitivity prompt setting, which tends to filter anomalies that are non-emergency. Table \ref{table:sample} shows the selected sample output of the LLM module.

\begin{table}[htbp]  
    \centering      
    \resizebox{0.6\linewidth}{!}{
    \begin{tabular}{ccc} 
        \toprule      
        Frame ID& Anomaly Index&Reason\\\midrule
 2290&0.85&'Car and people nearby.'\\
 3770& 0.2&'Green traffic light detected.'\\
 4430& 0.5&'Obstacles in path.'\\ 
  5450& 0.7& 'Car on the left'\\
  7000& 0& 'No immediate danger.'\\
 8230& 1&'Bike in close proximity'\\
  9800& 1& High risk of collision with multiple people\\\bottomrule
    \end{tabular}}
    \caption{Selected feedback and caption of the output of the LLM module. Frane ID indicates the video frame index, the Anomaly Index represents the predicted anomalies of LLM, and Reason represents the response message from LLM for anomaly interpretation.} 
    \label{table:sample}             
\end{table}

\subsection{Ablation Study}
To explore the contribution of different prompt modules, we conducted the ablation study of each module. 
Table \ref{table:abl} shows that the proposed system performs worse without specific prompt modules. For instance, while the instruction prompt is missing, the system may generate random content due to the confusion of the current task and lack of instruction. Moreover, missing region information of detected objects may also weaken the performance, as the system cannot evaluate the priority of the emergency.

\begin{table}[htbp]  
    \centering      
    \resizebox{0.4\linewidth}{!}{
    \begin{tabular}{ccccc} 
        \toprule      
        Sensitivity& Location &Instruction& AP&AUC\\\midrule
 Low&\cmark&\cmark&\textbf{88.01}&\textbf{87.26}\\ 
  Normal& \cmark& \cmark& 82.73&81.17\\
  High& \cmark& \cmark& 72.35&77.29\\
  Low& \cmark& \xmark& 68.84&73.64\\
  Low& \xmark& \cmark& 69.57&75.56\\
  \xmark& \cmark& \cmark& 69.16&80.39\\ \bottomrule
    \end{tabular}}
    \caption{Ablation study. \cmark  indicates incorporated modules of system, and \xmark  indicate missing modules} 
    \label{table:abl}             
\end{table}

Moreover, we find that LLM produced different performances with different sensitivity prompts. Unexpectedly, Low system sensitivity appears higher accuracy and precision, as the system tries to catch True Positive cases as much as possible and avoid false alarms. This is significant for visually impaired navigation, as the user can efficiently avoid misinformation and frequent-unnecessary alerts.


\subsection{Performance Evaluation}
We further explore the performance efficiency of the proposed system on multiple platforms to reveal its potential for other applications.

Latency: As shown in Table \ref{table:fps}, we measured end-to-end system latency and individual module processing times to identify bottlenecks and optimize for real-time performance. Results indicated an average end-to-end latency of $\textbf{60}$ \textbf{ms} on the mobile device (e.g., smartphone) with neural engines, ensuring timely feedback.

\begin{table}[htbp]  
    \centering      
        \resizebox{1\linewidth}{!}{
    \begin{tabular}{cccccccc} 
        \toprule      
        Task& Model  & Framework &Chipset &Architecture&FPS &Latency &w/ Frame Compensation\\
        \midrule
 Detection& Yolo-v8l& Pytorch &V100 &GPU&22.01&45 ms &102.56\\ 
 Detection&  Yolo-v8x & Pytorch &V100 &GPU&14.22&71 ms &70.11\\
 Segmentation& Yolo-v8x-seg& Pytorch &V100 &GPU&12.06&83 ms &59.68\\
 Detection&  Yolov8-World& Pytorch &V100 &GPU&20.12&50 ms &98.06\\
 Detection&  Yolov8x-World-v2& Pytorch &V100 &GPU&16.74&62 ms &76.88\\
 Detection&    Yolov8x-World-v2&  CoreML&M2&CPU& 5.01& 199  ms &N/A\\
 Detection &Yolov8x-World-v2 & CoreML&M2&Neural Engine&19.60&51 ms &N/A\\
 Detection& Yolov8x-World-v2& CoreML&A16-Bionic&CPU&1.26 &789 ms &N/A\\
  Detection& Yolov8x-World-v2& CoreML&A16-Bionic&Neural Engine&16.24&61 ms &N/A\\ \bottomrule
    \end{tabular}}
    \caption{Object Detector Test on multiple platforms. A16-Bionic processors are used in iPhone 14 pro max, and M2 processors are widely used in Vision Pro and the latest Mac models. The PyTorch-based implementation was run on NVIDIA GPU.} 
    \label{table:fps}             
\end{table}

Economy: We further investigated the system latency and token consumption for economy evaluation. We designed three different modes for users to choose from: 
\begin{itemize}
    \item Voice only: only output voice messages for emergency response, minimum latency.
    \item Annotation: output both anomaly index and reason for system testing and practical annotation.
    \item  Full: output full information in a structured JSON format. (See the original prompt for more information)
\end{itemize}

\begin{table}[htbp]  
    \centering      
     \resizebox{0.9\linewidth}{!}{
    \begin{tabular}{ccccccc} 
        \toprule      
        Mode &LLM &Latency 
& Completion Tokens&Prompt Tokens& Total Tokens& Charge (USD/day)\\\midrule
  Voice only&GPT3.5 &407 
&  35&573& 608&2.44\\ 
  Annotation&GPT3.5 &628 
&  48&573& 617&2.58\\
 Full& GPT3.5 &1818 & 176& 1195& 1371&13.53\\ \bottomrule
    \end{tabular}}
    \caption{Economy and latency test.} 
    \label{table:token}             
\end{table}

As shown in Table \ref{table:token}, we estimated the cost of our system with different modes. The prices are calculated with an average of $\textbf{2}$ hours of daily usage and are based on the chatGPT API pricing policy.

Original prompt:
\begin{itemize}
    \item Prompt\_format\_full: \textit{'Please organize your output into this format: \
{ "scene": quickly describe the current situation for blind user; \
  "key\_objects": quickly and roughly locate the key objects for blind user; \
  "anomaly\_checker": quickly diagnose if there is potential danger for a blind person; \
  "anomaly\_label": output 1 if there is an emergency, output 0 if not; \
  "anomaly\_index": object\_id, danger\_index, estimate a score from 0 to 1 about each objects that may cause danger; \
  "voice\_guide": the main output to instant alert the blind person for emergency.}'}
  \item Prompt\_format\_voice: \textit{'Please organize your output into this format: \
{ "voice\_guide": the main output to instantly alert the blind person for an emergency.}'}
  \item Prompt\_format\_annotation: \textit{'Please organize your output into this format: \
{ "anomaly\_score": predict a score from 0 to 1 to evaluate the emergency level; \
  "reason": explain your annotation reason within 10 words.}'}
    
\end{itemize}


\section{Conclusion}
This research demonstrates the significant potential of combining lightweight mobile object detection with large language models to enhance accessibility for visually impaired individuals. Our system successfully provides real-time scene descriptions and hazard alerts, achieving low latency and demonstrating the flexibility of prompt engineering for tailoring LLM output to this unique domain. Our experiments highlight the importance of balancing detection accuracy with computational efficiency for mobile deployment. Prompt design is a key component of our system in guiding LLM responses and ensuring the relevance of generated descriptions. Additionally, the integration of user feedback proved invaluable for refining the system's usability and overall user experience.

While this project offers a promising foundation, further research is warranted. Explorations into even more advanced prompt engineering for complex scenarios would pave the way for the wide adoption of such assistive technologies. Our findings illustrate the power of integrating computer vision and large language models, leading to greater independence and safety in daily life: a true testament to AI's ability to improve the quality of life for all.

\bibliography{references} 
\bibliographystyle{unsrt}

\appendix \section{Original Prompt} 
This section illustrates all used prompts in the proposed system.

\subsection{LLM Instruction}
System instructions are usually directly fed into LLMs as a self-prompt and generally do not consume token usage.

Main Instruction:  \texttt{"You are a voice assistant for a visually impaired user, \
the input is the actual data collected by a phone camera, and the phone is always facing front, \
please provide the key information for the blind user to help him navigate and avoid potential danger. \
Please note that the center\_x and center\_y represent the object location (proportional to the image), \
object height and width are also a proportion."
}
System Sensitivity Prompt:  \texttt{"System sensitivity: Incorporate the sensitivity setting in your response. \
For a low-sensitivity setting, identify and report only imminent and direct threats to safety. \
For medium sensitivity, include potential hazards that could pose a risk if not avoided. \
For high sensitivity, report all detected objects that could cause any inconvenience or danger.\
Current sensitivity: low."}

Location Prompt: \texttt{"The location information \(center_x, center_y, height, width\) of objects is the proportion to the image, the detected objects are categorized into 4 type based on the image region. Left and Right: objects located on left 25\% or right 25\% of the image, these objects are usually moving and has large proportion.Front: objects that may still far away, can be used to discriminate the current situation.Ground: objects that may nearby."}

Motion Prompt: \texttt{"{Using the information from last frame and current frame to analyze the movement (speed and direction) \
and location of each object to determine its trajectory relative to the user.\
Use this information to assess whether an object is moving towards the user or they are static. \
If moving, how quickly a potential collision might occur based on the object's speed and direction of movement.}"}

\subsection{LLM Prompt}
LLM prompts are the master prompts that are directly input from the user end, text tokens are counted to the usage. To control and optimize the usage, we designed three different output modes. Furthermore, the designed prompt will guide LLM to generate a structured data format (dictionary, list, JSON, etc.).

Full diagnose mode:  \texttt{'Please organize your output into this format: \
{ "scene": quickly describe the current situation for blind user; \
  "key\_objects": quickly and roughly locate the key objects for blind user; \
  "anomaly\_checker": quickly diagnose if there is potential danger for a blind person; \
  "anomaly\_label": output 1 if there is an emergency, output 0 if not; \
  "anomaly\_index": object\_id, danger\_index, estimate a score from 0 to 1 about each objects that may cause danger; \
  "voice\_guide": the main output to instant alert the blind person for emergency.}'}
  
Voice-only mode:  \texttt{'Please organize your output into this format: \
{ "voice\_guide": the main output to instantly alert the blind person for an emergency.}'}

Annotation mode:  \texttt{'Please organize your output into this format: \
{ "anomaly\_score": predict a score from 0 to 1 to evaluate the emergency level; \
  "reason": explain your annotation reason within 10 words.}'}

\subsection{Other Prompts}

Detection Classes Switch: \texttt{"The user is switching the scene to {custom\_scene} please generate a new list that contains the top 100 related objects, including especially road hazards and possible obstacles"}

Interest Target Setting: \texttt{"Please analyze the user command and extract the user required object, output into this format: {"add": object\_name}."}

\section{Detection Labels for Custom Scenes} 
This section illustrates the detection classes generated by GPT-4 that are customized for specific scenes.

Visually impaired navigation: \texttt{[ 'car', 'person', 'bus', 'bicycle', 'motorcycle', 'traffic light', 'stop sign',
    'fountain', 'crosswalk', 'sidewalk', 'door', 'stair', 'escalator', 'elevator', 'ramp',
    'bench', 'trash can', 'pole', 'fence', 'tree', 'dog', 'cat', 'bird', 'parking meter',
    'mailbox', 'manhole', 'puddle', 'construction sign', 'construction barrier',
    'scaffolding', 'hole', 'crack', 'speed bump', 'curb', 'guardrail', 'traffic cone',
    'traffic barrel', 'pedestrian signal', 'street sign', 'fire hydrant', 'lamp post',
    'bench', 'picnic table', 'public restroom', 'fountain', 'statue', 'monument',
    'directional sign', 'information sign', 'map', 'emergency exit', 'no smoking sign',
    'wet floor sign', 'closed sign', 'open sign', 'entrance sign', 'exit sign',
    'stairs sign', 'escalator sign', 'elevator sign', 'restroom sign', 'men restroom sign',
    'women restroom sign', 'unisex restroom sign', 'baby changing station',
    'wheelchair accessible sign', 'braille sign', 'audio signal device', 'tactile paving',
    'detectable warning surface', 'guide rail', 'handrail', 'turnstile', 'gate',
    'ticket barrier', 'security checkpoint', 'metal detector', 'baggage claim',
    'lost and found', 'information desk', 'meeting point', 'waiting area', 'seating area',
    'boarding area', 'disembarking area', 'charging station', 'water dispenser',
    'vending machine', 'ATM', 'kiosk', 'public telephone', 'public Wi-Fi hotspot',
    'emergency phone', 'first aid station', 'defibrillator',
    'tree', 'pole', 'lamp post', 'staff', 'road hazard']}

Urban Walking: \texttt{['pedestrian', 'cyclist', 'car', 'bus', 'motorcycle', 'scooter', 'electric scooter',
    'traffic light', 'stop sign', 'crosswalk', 'sidewalk', 'curb', 'ramp', 'stair', 'escalator', 
    'elevator', 'bench', 'trash can', 'pole', 'fence', 'tree', 'fire hydrant', 'lamp post',
    'construction barrier', 'construction sign', 'scaffolding', 'hole', 'crack', 'speed bump', 
    'puddle', 'manhole', 'drain', 'grate', 'loose gravel', 'ice patch', 'snow pile', 'leaf pile',
    'standing water', 'mud', 'sand', 'street sign', 'directional sign', 'information sign',
    'parking meter', 'mailbox', 'bicycle rack', 'outdoor seating', 'planter box', 'bollard', 
    'guardrail', 'traffic cone', 'traffic barrel', 'pedestrian signal', 'crowd', 'animal', 'dog', 
    'bird', 'cat', 'public restroom', 'fountain', 'statue', 'monument', 'picnic table', 
    'outdoor advertisement', 'vendor cart', 'food truck', 'emergency exit', 'no smoking sign', 
    'wet floor sign', 'closed sign', 'open sign', 'entrance sign', 'exit sign', 'stairs sign', 
    'escalator sign', 'elevator sign', 'restroom sign', 'braille sign', 'audio signal device', 
    'tactile paving', 'detectable warning surface', 'guide rail', 'handrail', 'turnstile', 
    'gate', 'security checkpoint', 'water dispenser', 'vending machine', 'ATM', 'kiosk',
    'public telephone', 'public Wi-Fi hotspot', 'emergency phone', 'charging station',
    'first aid station', 'defibrillator', 'tree', 'pole', 'lamp post', 'staff', 'road hazard']}

Walking General: \texttt{['vehicles',  
    'pedestrians',  
    'traffic signs and signals', 
    'roadway features', 
    'surface conditions', 
    'street furniture', 
    'construction areas', 
    'vegetation', 
    'animals',  
    'public amenities',  
    'navigation aids',  
    'temporary obstacles', 
    'emergency facilities',  
    'transportation hubs',  
    'electronic devices',
    'safety features']}

Urban Walking Hazards: \texttt{['person', 'cyclist', 'car', 'bus', 'motorcycle', 'scooter', 'fountain',
    'red traffic light', 'green traffic light',
    'stop sign', 'curb', 'ramp', 'stair', 'escalator',
    'elevator', 'bench', 'trash can', 'pole', 'tree', 'fire hydrant', 'lamp post',
    'construction barrier', 'construction sign', 'scaffolding', 'hole', 'crack', 'speed bump',
    'puddle', 'manhole', 'drain', 'grate', 'loose gravel', 'ice patch', 'snow pile', 'leaf pile',
    'standing water', 'mud', 'sand', 'street sign', 'directional sign', 'information sign',
    'parking meter', 'mailbox', 'bicycle rack', 'outdoor seating', 'planter box', 'bollard',
    'guardrail', 'traffic cone', 'traffic barrel', 'pedestrian signal', 'crowd', 'animal', 'dog',
    'bird', 'cat', 'public restroom', 'fountain', 'statue', 'monument', 'picnic table',
    'outdoor advertisement', 'vendor cart', 'food truck', 'emergency exit', 'no smoking sign',
    'wet floor sign', 'closed sign', 'open sign', 'entrance sign', 'exit sign', 'stairs sign',
    'escalator sign', 'elevator sign', 'restroom sign', 'braille sign', 'audio signal device',
    'tactile paving', 'detectable warning surface', 'guide rail', 'handrail', 'turnstile',
    'gate', 'security checkpoint', 'water dispenser', 'vending machine', 'ATM', 'kiosk',
    'public telephone', 'emergency phone', 'charging station',
    'first aid station', 'defibrillator', 
    'oil spill', 'road debris', 'branches', 'water'
    'low-hanging signage', 'road signs', 'roadworks', 'excavation sites', 'utility works',
    'fallen objects', 'spilled cargo', 'flood', 'ice', 'snowdrift', 'landslide debris',
    'erosion damage', 'parked vehicles', 'moving equipment',
    'large gatherings', 'parade', 'marathon', 'street fair',
    'scaffolding',
    'electrical hazards', 'wire tangle', 'manhole covers', 'street elements',
    'road hazards', 'toxic spill', 'biohazard materials',
    'wildlife crossings', 'stray animals', 'pets', 'flying debris', 'air pollution','smoke plumes', 'dust storms', 'sandstorms', 'floods', 'road crack']}

Walking test:  \texttt{[
    'vehicle', 
    'pedestrian', 
    'cyclist',
    'traffic signal',  
    'street sign',
    'crosswalk',
    'sidewalk',
    'curb',
    'ramp',
    'stair',
    'escalator',
    'elevator',
    'public seating',
    'trash receptacle', 
    'street furniture',  
    'tree',
    'construction site', 
    'road obstruction', 
    'loose materials', 
    'slick surface', 
    'animal', 
    'outdoor advertisement',
    'vendor',  
    'water feature',  
    'monument',  
    'information point', 
    'access point',  
    'safety equipment',  
    'navigation aid', 
    'public amenity', 
    'transport hub',  
    'obstacle crowd'
]}

Annotation mask = \texttt{{'people', 'human face', 'car license plate', 'license plate', 'plate'}}

\end{document}